%% file: neurips_2026.tex
\documentclass{article}

 \usepackage[preprint]{neurips_2026}

\usepackage[utf8]{inputenc} 
\usepackage[T1]{fontenc}    
\usepackage{hyperref}       
\usepackage{url}            
\usepackage{booktabs}       
\usepackage{amsfonts}       
\usepackage{nicefrac}       
\usepackage{microtype}      
\usepackage{xcolor}         

\usepackage{multirow}
\usepackage{array}
\usepackage{graphicx}
\usepackage{amsmath,amssymb}
\usepackage{hyperref}
\usepackage{enumitem}
\usepackage{amsfonts}
\usepackage{listings}
\usepackage[most]{tcolorbox}
\usepackage{colortbl}
\newtheorem{definition}{Definition}

\usepackage{wrapfig}

\hypersetup{colorlinks=true, linkcolor=blue, citecolor=blue, urlcolor=blue}
\definecolor{gain}{RGB}{200,50,50}      
\definecolor{drop}{RGB}{50,150,50}                 
\newcommand{\gup}[1]{\textcolor{gain}{#1}}
\newcommand{\gdn}[1]{\textcolor{drop}{#1}}          
\definecolor{promptbg}{HTML}{F7FAFC}
\definecolor{promptframe}{HTML}{2F5D7E}
\definecolor{prompttitlebg}{HTML}{2F5D7E}
\definecolor{prompttitlefg}{HTML}{FFFFFF}

\lstdefinestyle{promptstyle}{
    basicstyle=\ttfamily\footnotesize,
    breaklines=true,
    breakatwhitespace=true,
    columns=fullflexible,
    keepspaces=true,
    showstringspaces=false
}

  \tcbset{
    stageiibox/.style={
      width=\columnwidth,
      top=6pt,
      bottom=6pt,
      left=8pt,
      right=8pt,
      boxsep=2pt,
      arc=2pt,
      colback=black!2!white,
      colframe=black!55,
      coltitle=white,
      colbacktitle=black!80,
      fonttitle=\bfseries,
      enhanced,
      breakable,
      attach boxed title to top left={yshift=-0.08in,xshift=0.12in},
      boxed title style={
        boxrule=0pt,
        arc=1pt,
        colframe=white,
        colback=black!80,
      },
    }
  }
  \newtcolorbox{AIbox}[2][]{stageiibox,title=#2,#1}

\newcommand{\best}[1]{\cellcolor{gray!25}\textbf{#1}}
\newcommand{\second}[1]{\cellcolor{gray!10}\textbf{#1}}
\title{MolEmb: Multimodal Large Language Models Can Be Strong Molecular Embedding Models}

\author{%
  Xinjian Zhao $\S$ \P, Xiangru Jian $\diamondsuit$, Yaoyao Xu $\S$ \P, Xiaozhuang Song $\S$ \P, Wei Pang $\S$, \\\textbf{Lei Bai \P, Tianshu Yu $\S$ \P} \\
\textsuperscript{$\S$} The Chinese University of Hong Kong, Shenzhen \\
\textsuperscript{\P} Shanghai Artificial Intelligence Laboratory\\
\textsuperscript{$\diamondsuit$}  Cheriton School of Computer Science, University of Waterloo\\
 \texttt{xinjianzhao1@link.cuhk.edu.cn}, \texttt{xiangru.jian@uwaterloo.ca},
 \\
 \texttt{\{yaoyaoxu,xiaozhuangsong1,weipang\}@link.cuhk.edu.cn},
 \\ \texttt{leibai@pjlab.org.cn},
 \texttt{yutianshu@cuhk.edu.cn} 
}

\begin{document}

\maketitle

\begin{abstract}
Molecular embedding models can serve as foundational infrastructure for computational chemistry and drug discovery, where reusable vector representations support property prediction, virtual screening, and retrieval. Most molecular encoders are specialist models built around a single molecular view, producing unconditional vectors with no language interface for varying the representation. We ask whether multimodal large language models (MLLMs), which natively process images, text, and symbolic inputs, can instead serve as \emph{general molecular embedding models} that produce embeddings conditioned on both a molecular profile and a natural-language semantic context. We introduce \textbf{MolEmb}, a lightweight framework that adapts MLLMs by aligning molecular profiles with textual descriptions in a shared embedding space using a bidirectional contrastive objective. The resulting embedding model is competitive on molecular property prediction and supports cross-modal molecule--text retrieval in the same space. We further introduce \textbf{MolCAR}, a diagnostic benchmark for context-aware retrieval, and find that context-aware molecular embedding is primarily a data property of the supervision. These results suggest that MLLMs are not merely chemistry assistants or generators, but a viable and extensible route to general molecular embedding models.

\end{abstract}

\input{main}

\clearpage
\bibliographystyle{plain}
\bibliography{ref}







\appendix
\input{Appendix.tex}




\end{document}

%% file: main.tex
\section{Introduction}\label{sec:intro}

Molecular structure shapes physicochemical properties and interactions, from solubility and reactivity to binding affinity and toxicity~\cite{wildman1999prediction,wu2018moleculenet,sandfort2020structure,bian2026deepinteractions,deng2023systematic}. As deep learning becomes increasingly central to molecular modeling, learned molecular representations have become a common interface for property prediction, virtual screening, similarity search, and retrieval-augmented scientific reasoning~\cite{gilmer2017neural,cao2024large,ying2026neural,zhong2025benchmarking}. Molecular embedding models can serve as foundational infrastructure for these workflows. We use this term for models that provide stable, reusable embeddings, in the spirit of embedding models in natural language processing and information retrieval, rather than only an intermediate state of a task-specific prediction pipeline.

The field has produced such embeddings primarily through specialist encoders over molecular graphs, SMILES strings, or molecular geometry~\cite{edwards2022translation,zhou2023uni,bian2026deep}. These encoders are highly effective, especially for molecular prediction, but their representation interfaces are typically fixed in advance: one molecular view is encoded into a single unconditional vector. 
This raises a different embedding-model question: can molecular representations be exposed through a reusable interface whose input can vary by molecular view and semantic context?

To make this question precise, we study what we call a \emph{general molecular embedding model}. 
Here, \emph{general} emphasizes two properties beyond the usual reusable-vector view of embedding models. First, multi-view input: molecules can be described through complementary forms, such as 2D depictions, SMILES strings, conformers, and textual annotations, and a flexible model should accept the views available in a given workflow. Second, semantic conditioning: scientific workflows often impose different task lenses and domain expectations on the same molecule; a toxicity-focused query should emphasize different molecular evidence than a solubility-focused one. We formalize this object in Section~\ref{sec:general-encoder}.

MLLMs~\cite{achiam2023gpt,bai2025qwen3,comanici2025gemini} are natural candidates for such models. They process heterogeneous inputs such as images, text, and symbolic inputs; they can receive semantic context through natural language; and their hidden states can be pooled into fixed-length embeddings. This structural fit is timely because the broader foundation-model community has begun to treat generative backbones as embedding models. LLM2Vec~\citep{behnamghader2024llm2vec} and VLM2Vec~\citep{jiang2024vlm2vec} show that generative backbones can be adapted for retrieval and representation learning. This shift is also visible in industry-scale foundation-model systems: model families such as Qwen3-VL-Embedding~\citep{li2026qwen3} and Gemini Embedding~\citep{lee2025gemini} expose embedding-oriented variants, and DeepSeek-OCR 2 explores LM-style modules as visual encoders~\citep{wei2026deepseek}. Together, these developments suggest that foundation models are increasingly being used not only as generators, but also as reusable representation backbones.

In chemistry and broader scientific applications of MLLMs, however, the dominant framing remains generative: 
explanation, captioning, multimodal reasoning, or direct prediction~\cite{li2025chemvlm,tan2025chemmllm,lu2024moleculeqa,li2025rxnbench,zhou2025scientists}. These capabilities are valuable, but they leave underexplored the stable vector interface that many molecular workflows require. We therefore ask whether MLLMs can serve not only as chemistry assistants or generators, but as the basis of general molecular embedding models.

\begin{figure}[h]
    \centering
    \includegraphics[width=1\linewidth]{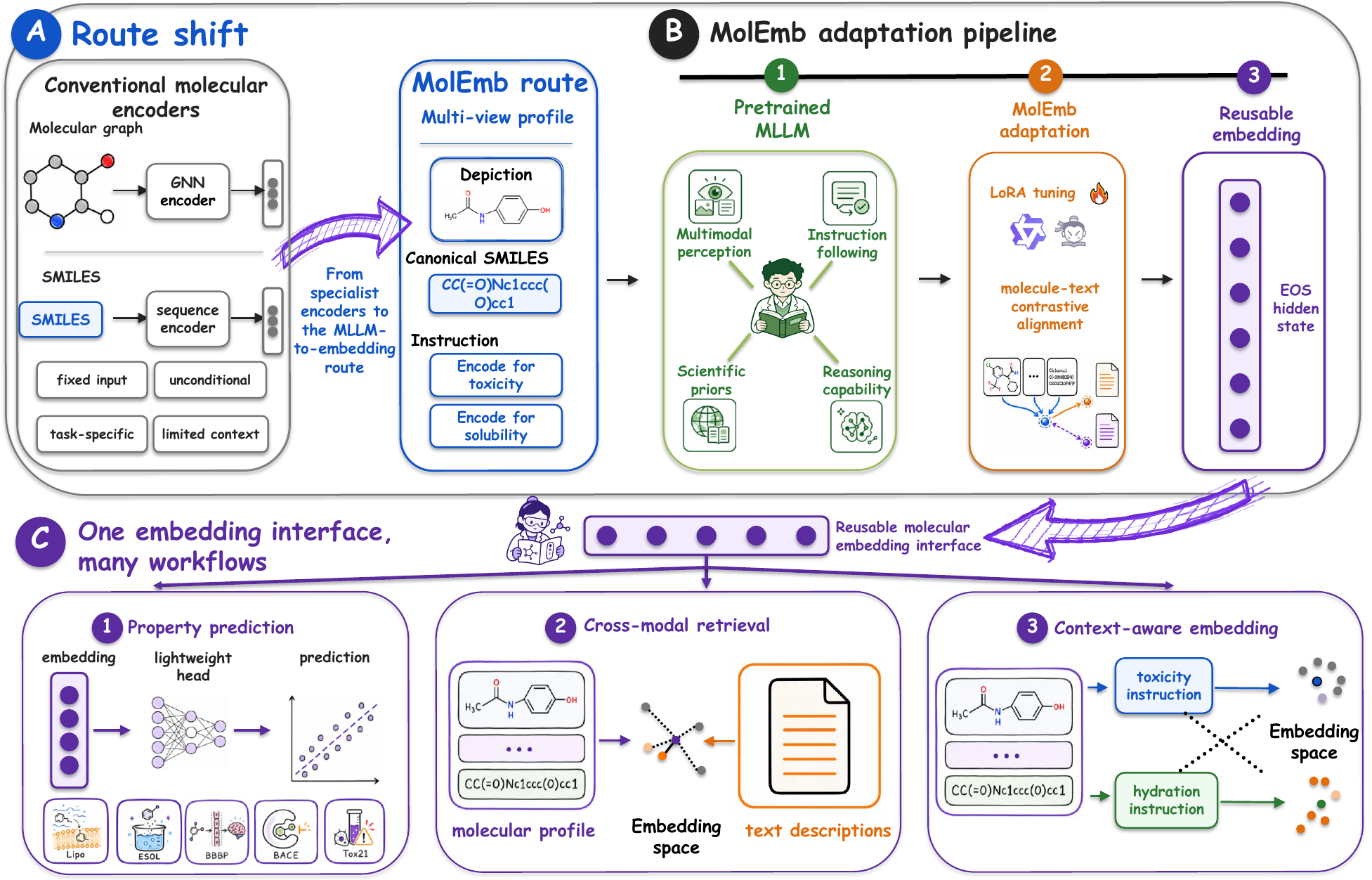}
    \caption{From specialist encoders to general molecular embedding models.
(A) The MLLM-to-embedding route shifts molecular representation from
fixed-view specialist encoders to an MLLM-based embedding interface that
accepts multi-view molecular profiles and natural-language task instructions.
(B) MolEmb instantiates this route by adapting a pretrained MLLM with
lightweight LoRA modules through molecule--text contrastive alignment; the EOS hidden state is used as the molecular embedding.
(C) The resulting reusable embedding interface supports property prediction,
cross-modal molecule--text retrieval, and context-aware embedding by varying
the task instruction.}
\label{fig:mainfig}
\end{figure}

We study this question through \emph{MolEmb}, a lightweight framework for adapting MLLMs into molecular embedding models. Each molecule is represented by a multi-view profile consisting of a 2D depiction and a canonical SMILES string, while semantic context is supplied through a text instruction when needed. We extract a fixed-length representation from the MLLM and align molecular profiles with textual descriptions in a shared embedding space. 
For molecular learning, this route is attractive because it allows molecular embedding models to share the backbone and interface of modern foundation models, and thus to benefit in principle from progress in multimodal pretraining, scientific-domain data, inference infrastructure, and embedding-oriented training. Figure~\ref{fig:mainfig} summarizes this route and the embedding-centered workflows evaluated below.

We evaluate this route along three axes. First, directly adapted MLLMs provide competitive representations for \emph{molecular prediction} across regression and classification benchmarks. Second, molecule--text contrastive alignment turns the same backbone into a reusable embedding model for \emph{cross-modal retrieval}, showing that the representation can support embedding-based matching rather than only prediction. Third, we introduce \textbf{MolCAR} for diagnosing context-aware retrieval, and use MolCAR-Train to show that task-structured supervision can induce context-aware structure in the embedding space.
Together, these results suggest that MLLMs are not merely chemistry assistants or generators, but a viable and extensible route to general molecular embedding models.

Our contributions are:
\begin{itemize}[leftmargin=*]
    \item We frame molecular representation learning through the lens of \emph{molecular embedding models}, emphasizing embeddings as reusable deliverables rather than only intermediate states of prediction pipelines, and formalize the target object as a \emph{general molecular embedding model}.
    \item We instantiate this perspective with MolEmb, a lightweight MLLM-based framework that 
    aligns multi-view molecular profiles with textual descriptions in a shared embedding space to produce reusable molecular embeddings.
    \item We evaluate the MLLM-to-embedding route across molecular prediction, cross-modal retrieval, and context-aware retrieval, showing both its practical viability and that context-aware molecular embedding is primarily a data property of the supervision.
\end{itemize}

\section{Related Work}

\textbf{Molecular Representation Learning.}
Molecular representation learning has been widely studied through graph-, sequence-, and geometry-based encoders. Graph neural networks encode atoms and bonds directly and have been strengthened by self-supervised objectives such as attribute masking, context prediction, and graph contrastive learning \citep{hu2019strategies,you2020graph,wang2022molecular,lin2022spectral,zhu2024molecular}. Sequence-based molecular language models instead treat SMILES strings as chemical text, enabling scalable Transformer pretraining with masked-language or sequence-to-sequence objectives \citep{chithrananda2020chemberta,ahmad2022chemberta,edwards2022translation}. A complementary line incorporates molecular geometry, using conformers, distances, or coordinate denoising to capture spatial molecular structure \citep{liu2021pre,zhou2023uni}. Beyond these molecular views, recent work also explores vision-based encoders that learn from rendered molecular diagrams or graph visualizations, showing that visual representations can capture useful structural and chemical patterns~\citep{cheng2026remol,zhao2026underappreciated}. These specialist encoders achieve strong property prediction performance, but they are usually tied to a fixed molecular view and produce unconditional embeddings, making it difficult to condition representations on natural-language task semantics.

\textbf{From Generative Models to Embedding Models.}
Recent work increasingly suggests that strong generative backbones can be repurposed as high-quality embedding models~\cite{behnamghader2024llm2vec,jiang2024vlm2vec,zhang2025qwen3,li2026qwen3,lee2025gemini,gunther2025jina,behnamghader2026llm2vec}. In the text-only setting, methods such as LLM2Vec and NV-Embed show that decoder-style language models can be adapted into competitive text embedding models for retrieval and representation learning~\cite{behnamghader2024llm2vec}. In the multimodal setting, VLM2Vec similarly shows that vision-language models can be converted into instruction-guided embedding models for multimodal retrieval~\cite{jiang2024vlm2vec}. This trend is also reflected in commercial model families, where dedicated embedding variants have emerged alongside generative backbones, such as the Qwen Embedding series~\cite{zhang2025qwen3,li2026qwen3} and Google’s Gemini embedding models~\cite{lee2025gemini}. These results suggest that generation and embedding should not be treated as disjoint model families: a strong generative backbone can often serve as the basis of a strong embedding model after suitable adaptation. 

More related work on multimodal foundation models in chemistry and broader scientific domains is discussed in Appendix~\ref{app:related-work-mfm}.

\section{General Molecular Embedding Models}
\label{sec:general-encoder}
\subsection{Definition}

Standard molecular encoders typically map a molecule, under a fixed input representation, to a single unconditional vector. This abstraction has been highly effective for specialist prediction tasks, but it leaves two aspects of molecular representation implicit. First, molecules are naturally multi-view objects: the same underlying molecule may be observed through a 2D depiction, a symbolic string, or other complementary annotations. Second, the most useful representation of a molecule need not be unconditional: the same molecule may require different embeddings under different semantic contexts.
Motivated by this perspective, we study what we call a \emph{general molecular embedding model}, which maps a molecular profile together with a semantic context to a vector representation in a shared embedding space.

\begin{definition}[General Molecular Embedding Model]
\label{def:general-embedding-model}
Let $\mathcal{M}$ be a set of molecules, identified up to a chosen
canonicalization scheme. Let $K$ be the number of possible molecular view types.
For each view type $k$, let $\mathcal{V}_k$ denote the corresponding view space,
such as 2D depictions, symbolic strings, conformers, or other molecule-specific
annotations. We define the molecular profile space as:
\begin{equation}
\mathcal{V}
=
\prod_{k=1}^{K} \bigl(\mathcal{V}_k \cup \{\varnothing_k\}\bigr).
\end{equation}
where $\varnothing_k$ denotes that view type $k$ is unavailable.
A molecular profile map $\mathcal{P}: \mathcal{M} \to \mathcal{V}$
assigns to each molecule $m \in \mathcal{M}$ its available views
$\mathcal{P}(m)$. Let $\mathcal{C}$ be a set of admissible semantic contexts,
and let $d$ be the embedding dimension. A \emph{general molecular embedding
model} is a parameterized map
$E_\theta : \mathcal{V} \times \mathcal{C} \to \mathbb{R}^d$ with
$\theta \in \Theta$, where $\Theta$ denotes the parameter space. For each
molecule $m \in \mathcal{M}$ and context $c \in \mathcal{C}$, we write
\begin{equation}
\mathbf{z}_c(m)
:=
E_\theta(\mathcal{P}(m), c)
\in \mathbb{R}^d .
\end{equation}
\end{definition}

\begin{definition}[Context-Aware Embedding Family]
\label{def:context-aware-family}
Given a general molecular embedding model $E_\theta$, the \emph{context-aware embedding family} of a molecule $m$ is
\[
\mathcal{Z}(m) := \{\mathbf{z}_c(m) : c \in \mathcal{C}\}.
\]
\end{definition}

Definitions~\ref{def:general-embedding-model}--\ref{def:context-aware-family} separate three roles that are often entangled in standard molecular representation learning. The molecular identity $m$ specifies the underlying molecule, the profile $\mathcal{P}(m)$ specifies the available observations of that molecule, and the context $c$ specifies the semantic lens under which the representation is formed. Multi-view input is captured through $\mathcal{P}(m)$, semantic conditioning through $c$, and a common downstream interface through the shared codomain $\mathbb{R}^d$. Under this formulation, a fixed molecule need not correspond to a single unconditional vector. Instead, it may induce a family of context-dependent embeddings $\mathcal{Z}(m)$, giving rise to a context-aware embedding geometry in which changing the semantic context may move the same molecule toward different regions of the shared space while preserving a common interface for downstream use.

Whether such an object can be realized in practice is an empirical question. Section~\ref{sec:experiments} tests whether multimodal large models instantiate it in a way that supports downstream prediction, retrieval, and context-aware behavior.
In this paper, we instantiate $\mathcal{P}(m)$ with a 2D depiction and a canonical SMILES string, and realize $c$ through natural-language instructions.

\subsection{MLLMs as a Natural Instantiation}
\label{sec:mllm-instantiation}
MLLMs are well-suited to instantiate
Definition~\ref{def:general-embedding-model} for both structural and strategic reasons. Structurally, they process heterogeneous token sequences, including image patches, text tokens, and symbolic strings. This allows the profile space $\mathcal{V}$ to include molecular depictions, SMILES, and other molecule-specific inputs, while allowing semantic contexts in $\mathcal{C}$ to be expressed through natural-language instructions. Strategically, the pretrained parameters of the MLLM provide a reusable backbone for $E_\theta$: improvements in multimodal pretraining, instruction following, and embedding-oriented adaptation can in principle be inherited without redesigning the molecular embedding interface. More broadly, the language grounding, instruction-following behavior, and broad scientific priors of such backbones make them natural candidates for interpreting semantic contexts in $\mathcal{C}$.

To instantiate $E_\theta$ with an MLLM, we construct an input sequence
\begin{equation}
    x(m,c) = \phi(\mathcal{P}(m), c),
\end{equation}
where $\phi$ formats the available molecular views and semantic context into the
model input. The MLLM defines a hidden-state map $\mathbf{H}_\theta : \mathcal{X} \to \mathbb{R}^{L \times d}$, where $\mathcal{X}$ is the input-token space and $L$ is the resulting sequence length. A pooling rule $\rho : \mathbb{R}^{L \times d} \to \mathbb{R}^d$ then maps the hidden states to a fixed-length vector:
\begin{equation}
E_\theta(\mathcal{P}(m), c)
:=
\rho\!\left(\mathbf{H}_\theta(x(m,c))\right)
\in \mathbb{R}^d.
\end{equation}

In our implementation, $\rho$ selects the hidden state at the end-of-sequence position. This gives a concrete instantiation of Definition~\ref{def:general-embedding-model} with a pretrained MLLM backbone. However, pretrained MLLM representations are not automatically aligned to molecular semantics. Section~\ref{sec:experiments} empirically examines this gap through molecule--text retrieval probes.

\subsection{MolEmb: Adapting MLLMs through Molecule--Text Alignment}
\label{sec:molemb}

MolEmb builds on the MLLM instantiation above and aligns molecular profiles with textual descriptions within a shared embedding space. Let $\theta_0 \in \Theta$ denote the pretrained MLLM parameters. We keep the pretrained backbone fixed and introduce a small trainable adapter for molecular adaptation. In our implementation, the adapter is parameter-efficient, with exact placement specified in the experimental setup. 

We optimize the trainable adapter using paired molecule-description data. During optimization, we write the current model parameters as $\theta$. For a molecule-description pair $(m_i,t_i)$ from an external corpus such as MolTextNet~\cite{zhu2025moltextnet} or ChEBI-20-MM~\cite{liu2025quantitative}, the molecular embedding is obtained as:
\begin{equation}
\mathbf{z}^{\mathrm{mol}}_i
=
E_{\theta}\bigl(\mathcal{P}(m_i), c_{\mathrm{enc}}\bigr),
\end{equation}
where $c_{\mathrm{enc}} \in \mathcal{C}$ is a fixed molecule-encoding instruction shared across all molecule--text alignment pairs. Let $\phi_{\mathrm{desc}}$ format a textual description into the MLLM input-token space. The description embedding is obtained by passing $t_i$ through the same backbone:
\begin{equation}
\mathbf{z}^{\mathrm{desc}}_i
=
\rho\!\left(\mathbf{H}_{\theta}(\phi_{\mathrm{desc}}(t_i))\right).
\end{equation}
Both embeddings lie in the shared space $\mathbb{R}^d$. We optimize a bidirectional contrastive objective:
\begin{equation}
\label{eq:contrastive}
\mathcal{L}
=
\frac{1}{2}\Bigl(
\mathcal{L}_{\mathrm{NCE}}
(\mathbf{z}^{\mathrm{mol}}, \mathbf{z}^{\mathrm{desc}})
+
\mathcal{L}_{\mathrm{NCE}}
(\mathbf{z}^{\mathrm{desc}}, \mathbf{z}^{\mathrm{mol}})
\Bigr),
\end{equation}
where the molecule-to-description direction is
\begin{equation}
\label{eq:nce-mol-desc}
\mathcal{L}_{\mathrm{NCE}}
(\mathbf{z}^{\mathrm{mol}}, \mathbf{z}^{\mathrm{desc}})
=
-\frac{1}{B}\sum_{i=1}^{B}
\log
\frac{
\exp\bigl(\mathrm{sim}(\mathbf{z}^{\mathrm{mol}}_i,
\mathbf{z}^{\mathrm{desc}}_i)/\tau\bigr)
}{
\sum_{j=1}^{B}
\exp\bigl(\mathrm{sim}(\mathbf{z}^{\mathrm{mol}}_i,
\mathbf{z}^{\mathrm{desc}}_j)/\tau\bigr)
}.
\end{equation}
Here $B$ is the batch size, $\tau$ is the temperature, and $\mathrm{sim}(\cdot,\cdot)$ is cosine similarity. The description-to-molecule direction is defined symmetrically. After optimization, we denote the resulting parameters by $\theta^\ast$ and the adapted embedding model by $E_{\theta^\ast}$.

MolEmb keeps the architecture unchanged beyond the chosen input views and alignment objective, allowing us to isolate the MLLM-to-embedding route itself. This design treats the pretrained MLLM as a multimodal and language-grounded backbone, whose instruction-following behavior and broad scientific priors can be exposed through lightweight molecular alignment to support molecular prediction, cross-modal retrieval, and context-aware embedding.

The aligned model supports property prediction using a lightweight task-specific head, cross-modal retrieval by ranking candidates in the shared embedding space $\mathbb{R}^d$, and context-aware embedding by varying $c \in \mathcal{C}$ to obtain different representations of the same molecule. To isolate the contribution of alignment, we also evaluate a direct adaptation baseline that bypasses molecule--text alignment and trains the MLLM directly on downstream prediction tasks.

\section{Experiments}\label{sec:experiments}
The three properties of a general molecular embedding model in Definition~\ref{def:general-embedding-model} suggest three corresponding empirical questions, which structure the rest of this section. (i) Is $E_\theta(\mathcal{P}(m), c)$ a useful predictor when reused across endpoints (\emph{molecular prediction}, Section~\ref{sec:prediction})? (ii) Is the codomain $\mathbb{R}^d$ a genuinely shared molecule--text space rather than a within-modality cluster (\emph{cross-modal retrieval}, Section~\ref{sec:retrieval})? (iii) Is the context-aware embedding family $\mathcal{Z}(m) = \{\mathbf{z}_c(m): c \in \mathcal{C}\}$ non-degenerate, in the sense that $c$ moves the embedding in task-meaningful directions for a fixed $m$ (\emph{context-aware embedding}, Section~\ref{sec:context})? Each subsection asks whether the MLLM-to-embedding route already supports the corresponding property, or whether targeted supervision is needed.

\subsection{Molecular Property Prediction}
\label{sec:prediction}

\textbf{Experimental Setup.}
We use eight standard molecular property benchmarks under scaffold splits~\cite{wu2018moleculenet,hu2020open}. The regression benchmarks are ESOL, Lipophilicity, and FreeSolv, and we report root mean squared error (RMSE). The classification benchmarks are BACE, BBBP, ClinTox, Tox21, and SIDER, and we report ROC-AUC. For multi-label datasets such as Tox21 and SIDER, ROC-AUC is macro-averaged over valid labels following the standard protocol. Dataset statistics and split sizes are provided in Appendix~\ref{app:benchmark-details}.
The main prediction table compares MolEmb with two groups of molecular baselines. The first group contains supervised graph neural networks, represented by S-GIN and R-GIN~\cite{lin2022spectral}. The second group contains molecular pretraining methods, including InfoGraph~\cite{sun2019infograph}, GraphCL~\cite{you2020graph}, MVGRL~\cite{hassani2020contrastive}, AD-GCL~\cite{suresh2021adversarial}, JOAO~\cite{you2021graph}, GCL-SPAN~\cite{lin2022spectral}, Mole-BERT~\cite{xia2023mole}, and MolTextNet~\cite{zhu2025moltextnet}. These baselines provide context for both task-specific supervised learning and specialist molecular representation learning.
For MolEmb, we evaluate three MLLM backbones: Intern-S1-mini (8B)~\cite{bai2025intern}, Qwen3-VL-8B~\cite{bai2025qwen3}, and Qwen3.5-0.8B~\cite{qwen35blog}. Each backbone is evaluated under three training conditions. \textbf{Direct} trains the pretrained MLLM on each downstream property task without prior molecule--text alignment. \textbf{MolTextNet} first aligns the embedding model on the MolTextNet molecule-description corpus and then fine-tunes it for downstream prediction. \textbf{Mixed} performs alignment on a mixture of molecule--text sources, broadening the semantic coverage of the alignment supervision through MolTextNet~\cite{zhu2025moltextnet}, KnowMol-100k~\cite{yang2025knowmol}, and ChEBI-20-MM~\cite{liu2025quantitative}. All three corpora use the same 98/1/1 train/valid/test split protocol; sizes are reported in Appendix Table~\ref{tab:alignment-corpora}. Results are reported as means and standard deviations over repeated runs with three seeds. Representative alignment examples are shown in Appendix~\ref{app:alignment-corpus-examples}, and implementation details are given in Appendix~\ref{app:implementation}.

Table~\ref{tab:main-results} compares MolEmb against representative supervised GNNs and molecular pretraining baselines. The prediction results test two basic requirements for the MLLM-to-embedding route: whether the interface can support molecular property learning, and whether broader molecule--text supervision improves downstream transfer.

\textbf{Direct adaptation provides a useful starting point.} Without any molecule--text alignment, MLLM-derived embedding models already reach the range of several pretraining-based GNN methods, particularly on regression tasks. This result does not by itself establish a reusable molecular embedding model, since the direct setting is still trained separately for each property task. Instead, it shows that the multimodal molecule interface can be optimized for molecular property learning before introducing molecule--text alignment. 
An ablation removing the 2D depiction (Fig.~\ref{fig:modality}, Appendix~\ref{app:modality-ablation}) further indicates that the image view contributes consistent though modest gains over a SMILES-only profile, supporting the multi-view input choice in Definition~\ref{def:general-embedding-model}.

\textbf{Alignment benefits depend on data diversity and coverage.} MolTextNet alignment improves several downstream tasks, showing that molecule-description matching can reshape the MLLM representation into a more useful molecular space. At the same time, MolTextNet is a single-source description corpus with limited stylistic and scientific coverage. The Mixed corpus is designed to broaden that supervision by adding KnowMol-100k and ChEBI-20-MM, which introduce names, functional groups, ontology-like definitions, and physicochemical cues. Its gains are endpoint-dependent rather than monotonic: Mixed alignment improves some tasks, notably ESOL and Tox21, while MolTextNet remains stronger on others. This pattern suggests both the value and the current limitation of molecule--text data: richer scientific coverage can help transfer, but generic description corpora still do not cover the full diversity of task-conditioned molecular knowledge.

\begin{table*}[t]
    \centering
    \caption{Main downstream comparison with representative molecular representation methods. Regression columns report RMSE, where lower is better; classification columns report ROC-AUC in percent, where higher is better. For each column, the \colorbox{gray!20}{\textbf{best}} and \second{second-best} results are highlighted.}
    \label{tab:main-results}
     \resizebox{\textwidth}{!}{
    \begin{tabular}{lcccccccc}
        \toprule
        & \multicolumn{3}{c}{\textbf{Regression (RMSE $\downarrow$)}} & \multicolumn{5}{c}{\textbf{Classification (ROC-AUC\% $\uparrow$)}} \\
        \cmidrule(lr){2-4} \cmidrule(lr){5-9}
        \textbf{Model} & \textbf{ESOL} & \textbf{Lipo} & \textbf{FreeSolv} & \textbf{BACE} & \textbf{BBBP} & \textbf{ClinTox} & \textbf{Tox21} & \textbf{SIDER} \\
        \midrule
        \rowcolor[RGB]{216, 216, 227}
        \multicolumn{9}{l}{\textit{Supervised GNN}} \\
        S-GIN & 1.173$\pm$0.057 & 0.757$\pm$0.018 & 2.755$\pm$0.349 & 72.97$\pm$4.00 & 68.17$\pm$1.48 & 88.14$\pm$2.51 & 74.91$\pm$0.51 & 57.60$\pm$1.40 \\
        R-GIN & 1.706$\pm$0.180 & 1.075$\pm$0.022 & 7.526$\pm$2.119 & 75.07$\pm$2.23 & 64.48$\pm$2.46 & 72.29$\pm$4.15 & 71.53$\pm$0.74 & 62.29$\pm$1.12 \\
        \midrule
        \rowcolor[RGB]{228, 234, 247}
        \multicolumn{9}{l}{\textit{Pre-training Methods}} \\
        InfoGraph & 1.344$\pm$0.178 & 1.005$\pm$0.023 & 10.005$\pm$4.819 & 74.74$\pm$3.64 & 66.33$\pm$2.79 & 64.50$\pm$5.32 & 69.74$\pm$0.57 & 60.54$\pm$0.90 \\
        GraphCL & 1.272$\pm$0.089 & 0.910$\pm$0.016 & 7.679$\pm$2.748 & 74.32$\pm$2.70 & 68.22$\pm$1.89 & 74.92$\pm$4.42 & 72.40$\pm$1.01 & 61.76$\pm$1.11 \\
        MVGRL & 1.433$\pm$0.145 & 0.962$\pm$0.036 & 9.024$\pm$1.982 & 74.20$\pm$2.31 & 67.24$\pm$1.39 & 73.84$\pm$4.25 & 70.48$\pm$0.83 & 61.94$\pm$0.94 \\
        AD-GCL & 1.217$\pm$0.087 & 0.842$\pm$0.028 & 5.150$\pm$0.624 & 76.37$\pm$2.03 & 68.24$\pm$1.47 & 80.77$\pm$3.92 & 71.42$\pm$0.73 & 63.19$\pm$0.95 \\
        JOAO & 1.285$\pm$0.121 & 0.865$\pm$0.032 & 5.131$\pm$0.722 & 74.43$\pm$1.94 & 67.62$\pm$1.29 & 78.21$\pm$4.12 & 71.83$\pm$0.92 & 62.73$\pm$0.92 \\
        GCL-SPAN & 1.218$\pm$0.052 & 0.802$\pm$0.019 & 4.531$\pm$0.463 & 76.74$\pm$2.02 & 69.59$\pm$1.34 & 80.28$\pm$2.42 & 72.83$\pm$0.62 & 64.87$\pm$0.88 \\
        Mole-BERT & 1.015$\pm$0.030 & \second{0.676$\pm$0.017} & -- & 80.80$\pm$1.40 & 71.90$\pm$1.60 & 78.90$\pm$3.00 & 76.80$\pm$0.50 & 62.80$\pm$1.10 \\
        MolTextNet & 1.145$\pm$0.070 & 0.753$\pm$0.008 & 2.357$\pm$0.106 & \best{84.70$\pm$0.10} & 70.40$\pm$2.40 & 90.00$\pm$0.20 & 75.20$\pm$0.30 & 64.00$\pm$3.10 \\
        \midrule
        \rowcolor[RGB]{230, 239, 241}
        \multicolumn{9}{l}{\textit{MolEmb (Ours)}} \\
        Intern-S1-mini (Direct) & 0.827$\pm$0.003 & 0.700$\pm$0.008 & \second{1.839$\pm$0.031} & 75.66$\pm$1.49 & 73.09$\pm$0.25 & \second{99.51$\pm$0.11} & 77.69$\pm$0.59 & 61.28$\pm$1.31 \\
        Intern-S1-mini (MolTextNet) & 0.802$\pm$0.001 & 0.678$\pm$0.006 & 1.886$\pm$0.034 & 81.27$\pm$0.17 & \best{75.00$\pm$0.18} & \best{99.54$\pm$0.06} & 77.86$\pm$0.74 & \second{65.44$\pm$0.52} \\
        Intern-S1-mini (Mixed) & \best{0.771$\pm$0.012} & \best{0.673$\pm$0.004} & 1.965$\pm$0.080 & 78.65$\pm$2.15 & \second{74.89$\pm$1.46} & 99.50$\pm$0.18 & \best{79.24$\pm$0.71} & 64.78$\pm$0.08 \\
        Qwen3-VL-8B (Direct) & 0.864$\pm$0.016 & 0.792$\pm$0.009 & 2.029$\pm$0.056 & 79.22$\pm$1.34 & 70.39$\pm$0.69 & 97.03$\pm$2.84 & 75.77$\pm$0.19 & 59.99$\pm$0.38 \\
        Qwen3-VL-8B (MolTextNet) & 0.810$\pm$0.017 & 0.699$\pm$0.002 & 2.036$\pm$0.043 & 75.03$\pm$0.90 & 72.48$\pm$0.80 & 99.05$\pm$0.51 & 77.37$\pm$0.15 & 63.64$\pm$0.07 \\
        Qwen3-VL-8B (Mixed) & 0.793$\pm$0.004 & 0.689$\pm$0.001 & 1.936$\pm$0.008 & 76.37$\pm$0.13 & 74.77$\pm$0.21 & 98.70$\pm$0.46 & \second{78.08$\pm$0.07} & \second{64.85$\pm$0.19} \\ 
     Qwen3.5-0.8B (Direct) & 0.958$\pm$0.013 & 0.796$\pm$0.008 & 2.447$\pm$0.175 & 78.39$\pm$0.94 & 69.06$\pm$1.76 & 98.97$\pm$0.02 & 73.98$\pm$0.73 & 59.15$\pm$2.11 \\
        Qwen3.5-0.8B (MolTextNet) & 0.825$\pm$0.038 & 0.696$\pm$0.016 & \best{1.834$\pm$0.139} & \second{80.97$\pm$0.79} & 71.58$\pm$2.89 & 99.14$\pm$0.43 & 76.19$\pm$0.87 & 62.14$\pm$0.18 \\
        Qwen3.5-0.8B (Mixed) & \second{0.781$\pm$0.016} & 0.719$\pm$0.011 & 1.857$\pm$0.084 & 78.47$\pm$2.63 & 72.79$\pm$1.29 & 99.30$\pm$0.19 & 77.45$\pm$0.25 & 64.39$\pm$1.54 \\
        \bottomrule
    \end{tabular}}\vspace{-3.9mm}
\end{table*}

\begin{table*}[h]
    \centering
    \caption{Molecule-text retrieval on MolTextNet \texttt{test} split. m2t denotes molecule-to-text retrieval, t2m text-to-molecule retrieval, and R@$K$ denotes recall at $K$. All values are reported as fractions.}      \resizebox{\textwidth}{!}{
    \label{tab:retrieval-probe}
    \begin{tabular}{lcccccc}
        \toprule
        Model & m2t R@1 & m2t R@5 & m2t R@10 & t2m R@1 & t2m R@5 & t2m R@10 \\
        \midrule
        Qwen3-VL-Embedding-8B & 0.0347 & 0.0838 & 0.1185 & 0.0306 & 0.0828 & 0.1229 \\
        Qwen3.5-0.8B + MolEmb (MolTextNet) & 0.8397 & \textbf{0.9818} & \textbf{0.9963} & 0.8458 & \textbf{0.9805} & \textbf{0.9926}\\
        Intern-S1-mini + MolEmb (MolTextNet) & \textbf{0.8539} & 0.9811 & 0.9926 & \textbf{0.8488} & 0.9785 & 0.9923 \\
        Qwen3-VL-8B + MolEmb (MolTextNet) & 0.7343 & 0.9549 & 0.9842 & 0.7502 & 0.9572 & 0.9828 \\
        \bottomrule
    \end{tabular}}\vspace{-3.5mm}
\end{table*}
\subsection{Cross-Modal Retrieval}
\label{sec:retrieval}
Prediction alone does not establish that the learned embedding is reusable beyond a single task head. A general molecular embedding model should also support embedding-based operations such as similarity search and retrieval-augmented workflows. Specialist molecular encoders do not natively support these workflows: their unconditional vectors do not share a space with text, so molecule--text retrieval requires bolting on a separate text tower. A unified MLLM backbone offers a single interface in which such a space can be learned; the retrieval probe below asks whether the codomain $\mathbb{R}^d$ in Definition~\ref{def:general-embedding-model} actually behaves as a shared molecule--text space. We evaluate molecule-to-text and text-to-molecule retrieval on held-out MolTextNet data; the candidate pool contains 2{,}970 molecule--text pairs, so the random R@1 baseline is about $0.034\%$.

\textbf{Generic multimodal embeddings collapse on molecules.} A general-purpose multimodal embedding model, Qwen3-VL-Embedding-8B, achieves only about 3\% R@1 in both retrieval directions, indicating that current off-the-shelf multimodal embedding models have not formed a sufficiently reliable chemical embedding space and therefore offer little support for retrieval in this domain. After molecule--text contrastive alignment on MolTextNet train split, all MolEmb backbones exceed 73\% R@1 in both directions, and Intern-S1-mini and Qwen3.5-0.8B exceed 83\%. Intern-S1-mini attains the best R@1 in both retrieval directions, consistent with its being the only one of the three backbones with explicit scientific-domain pretraining.

Beyond cross-modal retrieval, a reusable molecular embedding is meant to serve many scientific purposes, such as reaction planning, virtual screening, toxicity assessment, materials design, or scientific literature search, where the most useful representation of a molecule plausibly differs across uses. Such context-aware embedding is the foundational capability that separates a general molecular embedding model from a task-specific molecule--text retriever.

\subsection{Context-Aware Embedding with MolCAR}
\label{sec:context}
Here we focus on whether such context-aware embedding can be \emph{acquired} by an MLLM-based embedding model. To make the question testable, we work in a controlled setting that holds molecule identity fixed and varies only the task lens: across eight task families drawn from standard property-prediction benchmarks, a query about blood-brain barrier penetration, toxicity, or aqueous interaction may all refer to the same molecular structure but should retrieve different evidence. We introduce \textbf{MolCAR} (\textbf{Mol}ecular \textbf{C}ontext-\textbf{A}ware \textbf{R}etrieval), a benchmark family for testing whether task instructions act as a routing signal when molecule identity alone is insufficient. MolCAR simulates a multi-task scientific retrieval scenario over a candidate pool of synthetic context documents: each target document anchors a real experimental outcome from one of eight task families (e.g., BACE, BBBP, ESOL) to its molecular profile, so the same molecule appears with multiple valid documents that differ only by task lens. This construction lets us hold molecule identity fixed and ask whether the model routes the query to the right document under a given task instruction.

\textbf{MolCAR-Structured} deterministically constructs target cards from real task labels and molecular descriptors, pairing each held-out molecule with multiple task contexts and reporting both global document retrieval and within-molecule context ranking. The evaluation split contains 4,092 molecule--context pairs over 1,653 held-out molecules from eight task families. \textbf{MolCAR-Natural} replaces structured targets with Intern-S1 (235B)-generated scientific notes anchored to the same observed outcomes and molecular profiles, reducing template regularity while preserving ground-truth anchors. \textbf{MolCAR-Train} is built from the training splits of the same downstream task collection after removing benchmark molecules by canonical SMILES, providing task-conditioned supervision without molecule leakage; like the eval pools, it comes in matched Structured and Natural variants. Rather than testing transfer to unseen task families, MolCAR asks a controlled diagnostic question: whether the model can use task language to select the appropriate context for a held-out molecular profile. See Appendix~\ref{app:molcar-prompts} for construction details.

\begin{AIbox}[colback=green!2!white,colframe=green!40!black,colbacktitle=green!55!black]{MolCAR Structured vs Natural Example}
\small
\textbf{Molecule query (same for both variants).}

\texttt{<image>} Encode this molecule for Tox21 toxicity outcomes. SMILES: BrC(Br)Br

\medskip
\textbf{MolCAR-Structured target (matched retrieval target).}

Report family: safety liability\\
Task: Tox21 toxicity outcomes\\
Observed result: This molecule is positive in 0 of 12 observed Tox21 assay panel.\\
Molecular profile: MW 252.7; cLogP 2.45; HBD 0; HBA 0; formal charge 0

\medskip
\textbf{MolCAR-Structured hard negative (same molecule, wrong task).}

Report family: aqueous interaction\\
Task: hydration free energy\\
Observed result: Observed hydration free energy is -2.130 kcal/mol.\\
Molecular profile: MW 252.7; cLogP 2.45; HBD 0; HBA 0; formal charge 0

\medskip
\textbf{MolCAR-Natural target (matched retrieval target).}

The molecule exhibits no positive responses across the 12 Tox21 assay panel, suggesting a low potential for toxicity in the tested endpoints. Its molecular profile, characterized by a moderate molecular weight (252.7) and balanced lipophilicity (cLogP 2.45), indicates favorable physicochemical properties that may contribute to its non-toxic profile. The absence of hydrogen bond donors and acceptors, along with a neutral formal charge, further supports its stability and reduced reactivity in biological systems.

\medskip
\textbf{MolCAR-Natural hard negative (same molecule, wrong task).}

The compound exhibits a measured hydration free energy of -2.130 kcal/mol, indicating moderate hydrophilic character despite its neutral charge and lack of hydrogen bond donors or acceptors. With a molecular weight of 252.7 and a cLogP of 2.45, the molecule demonstrates balanced lipophilicity and aqueous solubility properties. These features suggest limited polar interactions with water, consistent with its structural profile.
\end{AIbox}

Table~\ref{tab:molcar_structured} reports MolCAR-Structured results. Random baselines are approximately $0.024\%$ for Doc.\ R@1, $0.060\%$ for Mol.\ R@1, and $40.4\%$ for Context~R@1 due to variable context counts per molecule. The primary metric is Context~R@1, which ranks candidate context documents \emph{within the same molecule} and therefore factors out molecule identification. Before continued alignment, task instructions provide only weak control over same-molecule context ranking: Intern-S1-mini obtains 44.4\% Context~R@1 with task instructions and 40.4\% without them, while Qwen3.5-0.8B obtains 41.1\% and 40.4\%. The small Intern-S1-mini gap suggests residual instruction sensitivity, but both backbones remain near the context-level baseline. This is the key failure mode of generic molecule--text alignment: it can make molecular profiles and molecular descriptions comparable, but it does not reliably make the semantic context $c$ separate different valid descriptions of the same molecule. This pattern is consistent with the structure of the alignment supervision: as Appendix~\ref{app:alignment-corpus-examples} illustrates, the descriptions in MolTextNet, KnowMol100k, and ChEBI-20-MM are templated and narrow in linguistic distribution, so during alignment pretraining the model rarely sees the same molecule under varying task lenses.

\textbf{Continued alignment expands the context-aware embedding family.}
We further train the alignment-pretrained models on MolCAR-Train using the same bidirectional contrastive objective, a stage we call continued alignment. After this stage, Context~R@1 with task instructions reaches 99.8\% for Intern-S1-mini and 69.4\% for Qwen3.5-0.8B, far above the 40.4\% context-level baseline. The gain shows that continued alignment makes the embedding model sensitive to the task instruction, rather than merely improving molecule-level retrieval. In the language of Definition~\ref{def:general-embedding-model}, the alignment-pretrained model behaves as if $\mathbf{z}_c(m) \approx \mathbf{z}_{c'}(m)$ across the tested MolCAR task lenses, so the observed context-aware embedding family $\mathcal{Z}(m)$ is nearly collapsed; continued alignment expands $\mathcal{Z}(m)$ along task-meaningful directions.

\textbf{Context-aware embedding also holds with generated scientific notes.}
Table~\ref{tab:molcar_natural} (Appendix~\ref{app:molcar_result}) reports MolCAR-Natural, where Intern-S1 (235B)-generated scientist-facing notes replace structured target cards, broadening the target-text distribution while preserving molecule identity, task label, and observed result. This setting is closer to realistic scientific retrieval, where relevant documents are not written in a fixed schema. After continued alignment, both backbones show large task-conditioned gains; Qwen3.5-0.8B reaches slightly higher Context~R@1 under MolCAR-Natural than MolCAR-Structured (72.0\% vs.\ 69.4\%).%

\textbf{Context-aware embedding is a data property.} The MolCAR pattern gives a clean embedding-level reading: generic molecule--text corpora teach an unconditional chemical manifold, where molecular profiles and descriptions become comparable but the same molecule under different task lenses is not reliably separated. MolCAR-Train supplies the missing supervision by pairing molecular profiles with outcome-grounded target texts under explicit task labels, teaching the embedding model which semantic directions should be activated by the instruction. A complementary geometric view in Appendix~\ref{app:tsne-prompt-probe} is consistent with this reading: task-wise separation in the embedding space emerges only after continued alignment, and only when task instructions are present. Context-aware molecular embedding is therefore primarily a data property of the supervision, induced here by task-diverse, outcome-grounded molecule--text supervision. This points to a data-centric path for building stronger molecular embedding models: curate supervision that reflects the scientific task lenses under which the embeddings will be used.

\begin{table}[t]
\centering
\caption{MolCAR-Structured results.
\textbf{Base}: alignment-pretrained on the Mixed corpus.
\textbf{+ Continued Alignment}: after continued alignment on MolCAR-Train.
\textbf{Inst}: query with task-specific instruction; \textbf{NoInst}: generic query without task specification.
$\Delta$ = Inst $-$ NoInst (percentage points).
\textbf{Mol.\ R@1}: correct molecule in top-1.
\textbf{Doc.\ R@1}: correct document (molecule + context) in top-1.
\textbf{Doc.\ MRR}: $100\times$ mean reciprocal rank of the correct document.
\textbf{Doc.\ R@5}: correct document in top-5.
\textbf{Context R@1}: correct context ranked first \emph{within} the same molecule.}
\label{tab:molcar_structured}
\small
\setlength{\tabcolsep}{4pt}
\begin{tabular}{ll|ccc|ccc}
\toprule
\textbf{Model} & \textbf{Metric} & \multicolumn{3}{c|}{\textbf{Base}} & \multicolumn{3}{c}{\textbf{+ Continued Alignment}} \\
\cmidrule(lr){3-5} \cmidrule(lr){6-8}
 & & \textbf{Inst (\%)} & \textbf{NoInst (\%)} & \textbf{$\Delta$ (pp)} & \textbf{Inst (\%)} & \textbf{NoInst (\%)} & \textbf{$\Delta$ (pp)} \\
\midrule
\multirow{5}{*}{Intern-S1-mini} & Mol.\ R@1 & 10.5 & 10.6 & \gdn{-0.1} & 24.9 & 16.0 & \gup{+8.9} \\
 & Doc.\ R@1 & 4.5 & 4.3 & \gup{+0.2} & 24.8 & 6.3 & \gup{+18.5} \\
 & Doc.\ MRR & 11.3 & 10.7 & \gup{+0.6} & 39.8 & 14.6 & \gup{+25.2} \\
 & Doc.\ R@5 & 16.6 & 15.3 & \gup{+1.3} & 57.4 & 21.1 & \gup{+36.3} \\
 & Context R@1 & 44.4 & 40.4 & \gup{+4.0} & \textbf{99.8} & 40.4 & \gup{\textbf{+59.4}} \\
\midrule
\multirow{5}{*}{Qwen3.5-0.8B} & Mol.\ R@1 & 4.1 & 4.3 & \gdn{-0.2} & 12.4 & 10.6 & \gup{+1.8} \\
 & Doc.\ R@1 & 1.8 & 1.7 & \gup{+0.1} & 7.8 & 4.0 & \gup{+3.8} \\
 & Doc.\ MRR & 5.0 & 5.0 & \gup{+0.0} & 17.4 & 9.2 & \gup{+8.2} \\
 & Doc.\ R@5 & 6.5 & 6.5 & \gup{+0.0} & 25.2 & 12.8 & \gup{+12.4} \\
 & Context R@1 & 41.1 & 40.4 & \gup{+0.7} & \textbf{69.4} & 40.4 & \gup{\textbf{+29.0}} \\
\bottomrule
\end{tabular}\vspace{-1.9mm}
\end{table}

\section{Conclusion}

We studied whether multimodal large language models can be adapted into reusable molecular embedding models. MolEmb takes a simple route: represent each molecule through an image depiction, SMILES, and a text context; extract a MLLM representation; and align molecule and text embeddings with a contrastive objective. Across property prediction, molecule--text retrieval, and MolCAR context-aware retrieval, the results show that the MLLM interface is a viable starting point, molecular alignment is necessary for embedding-based retrieval, and generic molecule-description alignment alone is insufficient for reliable task-conditioned routing.
The broader implication is data-centric. Current molecule--text corpora provide useful structural and physicochemical semantics, but their coverage is narrow relative to the task-conditioned scientific language needed for context-aware molecular retrieval. Building more diverse, outcome-grounded, and task-aware molecular text corpora may therefore be as important as changing backbone scale or architecture.

%% file: Appendix.tex
\clearpage
\section{Limitations}\label{app:limitations}
MolEmb is evaluated as an adaptation route rather than as a fully scaled molecular foundation model. Our molecular profiles use a 2D depiction and canonical SMILES, leaving richer views such as conformers, reactions, spectra, and assay metadata for future instantiations of the profile map  $\mathcal{P}(m)$. MolCAR is designed as a controlled diagnostic for context-aware embedding: it holds 
molecule identity fixed and varies task lenses within eight property-related task families, but it does not test transfer to unseen task families or open-ended scientific search. Finally, the results suggest that context-aware molecular embedding depends strongly on the supervision. The Mixed corpus broadens semantic coverage, but its scale and diversity remain small relative to industrial embedding pretraining, so broader context-aware embedding will require larger, more carefully curated molecule--text supervision.

\section{Additional Related Work on Multimodal Foundation Models for Science}
\label{app:related-work-mfm}
\textbf{Multimodal Foundation Models for Science.}
Large multimodal models have increasingly been applied to chemistry and broader scientific domains for question answering, captioning, multimodal reasoning, and cross-modal understanding~\cite{su2022molecular,liu2023multi,luo2023molfm,li2025chemvlm,tan2025chemmllm,yan2025position,zhao2026vision}. Within chemistry, models such as MolFM~\cite{luo2023molfm} and ChemVLM~\cite{li2025chemvlm} study multimodal learning over molecular structure, text, and related scientific signals. More broadly, scientific multimodal foundation models such as Intern-S1 extend this paradigm beyond chemistry-specific settings toward general scientific understanding~\cite{bai2025intern}. Together, these works highlight an important structural advantage: multimodal large models naturally accommodate heterogeneous inputs and language-based interaction. However, much of the current literature remains generation-centric, focusing on answering, captioning, or reasoning over scientific inputs. 
In contrast, our work studies such backbones as embedding models: the goal is not primarily to generate chemistry-flavored text, but to produce reusable molecular representations that support prediction, retrieval, and transfer.
\section{Benchmark Details}
\label{app:benchmark-details}

\subsection{Downstream Prediction Datasets}

Table~\ref{tab:dataset-stats} summarizes the eight downstream datasets used in
the main prediction evaluation. All datasets are taken from the OGB / MoleculeNet
benchmark suite and use the standard scaffold split.

\begin{table}[h]
\centering
\caption{Downstream prediction datasets. Train/valid/test sizes follow the OGB scaffold split.}
\label{tab:dataset-stats}
\begin{tabular}{llrrrl}
\toprule
\textbf{Dataset} & \textbf{Task} & \textbf{Train} & \textbf{Valid} & \textbf{Test} & \textbf{Metric} \\
\midrule
ESOL  & Regression & 902 & 113 & 113 & RMSE \\
Lipophilicity  & Regression & 3360 & 420 & 420 & RMSE \\
FreeSolv  & Regression & 513 & 64 & 64 & RMSE \\
BACE & Binary cls. & 1210 & 152 & 152 & ROC-AUC \\
BBBP & Binary cls. & 1631 & 204 & 204 & ROC-AUC \\
ClinTox & Binary cls. & 1181 & 148 & 148 & ROC-AUC \\
Tox21 & Multi-label cls. (12) & 6264 & 783 & 784 & ROC-AUC \\
SIDER & Multi-label cls. (27) & 1141 & 143 & 143 & ROC-AUC \\
\bottomrule
\end{tabular}
\end{table}

For multi-label classification (Tox21, SIDER), we report the macro-averaged
ROC-AUC across all active tasks, following the standard OGB evaluation protocol.
ClinTox has two clinical trial outcome labels; we report the averaged ROC-AUC
over both.

\subsection{MolCAR Retrieval Metrics}
\label{app:molcar-metrics}

The MolCAR evaluation uses five retrieval metrics, all computed over a fixed
candidate pool containing all evaluation documents (molecule--context pairs).
Queries are multimodal molecular profiles, optionally augmented with a task
instruction (Inst) or using a generic no-instruction variant (NoInst).

\begin{itemize}
  \item \textbf{Mol.~R@1}: fraction of queries for which the ground-truth
    molecule appears in rank 1. Because multiple documents share the same
    molecule, this measures molecule-level identification rather than exact
    document matching.
  \item \textbf{Doc.~R@1}: fraction of queries for which the ground-truth
    document (correct molecule \emph{and} correct task context) appears in rank 1.
  \item \textbf{Doc.~MRR}: mean reciprocal rank of the ground-truth document
    across all queries.
  \item \textbf{Doc.~R@5}: fraction of queries for which the ground-truth
    document appears in the top-5 results.
  \item \textbf{Context~R@1}: fraction of queries for which the ground-truth
    context is ranked \emph{first among all documents sharing the same molecule}.
    This metric isolates instruction sensitivity from molecule identification:
    it asks whether the model correctly routes the query to the right task
    view given that the molecule is already identified. A random baseline
    achieves $1/K$ where $K$ is the number of context types for that molecule.
    Because molecules have different numbers of available task contexts, the aggregate random baseline for Context~R@1 is approximately $40.4\%$ (1,653/4,092) rather than $12.5\%$.
\end{itemize}

Context~R@1 is the primary metric for the context-aware embedding evaluation because it directly measures whether the model uses the task instructions as a routing signal, independently of how well it retrieves the correct molecule.

\subsection{MolCAR Dataset Construction}
\label{app:molcar-construction}

\textbf{Evaluation split (MolCAR-Structured).} Molecules are held out at the
canonical-SMILES level: any molecule appearing in the training splits of the
eight downstream datasets is removed from the evaluation pool using exact
canonical-SMILES matching. The evaluation set contains 4,092 molecule--context
pairs over 1,653 unique held-out molecules drawn from the test splits of the
eight task families. Each molecule is paired with all available task contexts,
forming a dense multi-context evaluation structure.

\textbf{Training split (MolCAR-Train).} The continued-alignment corpus is
built from the training splits of the same eight datasets after removing all
evaluation molecules by canonical SMILES. Each training instance maps one
molecular profile to one task-conditioned textual target, providing
same-molecule, multi-context supervision when a molecule appears in more than
one dataset.

\textbf{MolCAR-Natural.} Structured context documents are replaced by
scientific notes generated by Intern-S1 (235B parameters). The generation is
anchored to the same ground-truth outcome and molecular profile as the
structured document. The generation backbone (235B) is distinct from the
MolEmb backbones (Intern-S1-mini, Qwen3.5-0.8B) evaluated in the paper,
which avoids confound between data generation and model evaluation.

\subsection{MolCAR Construction Instructions and Templates}
\label{app:molcar-prompts}

This subsection lists the templates and instructions used to construct MolCAR. Three
ingredients are involved: the task-instructed query format, the structured
target card schema, and the natural-language generation prompt for
MolCAR-Natural.

\paragraph{Task-instructed queries.} All MolCAR queries follow a single
template, with only the task name varying across the eight task families:

\begin{AIbox}[colback=red!2!white,colframe=red!40!black,colbacktitle=red!55!black]{MolCAR Query Template}
\small
\texttt{<image>}\\
Encode this molecule for \textit{\{task\_name\}}. SMILES: \textit{\{smiles\}}
\end{AIbox}

The eight task names are:
\textit{BACE-1 inhibition},
\textit{blood-brain barrier penetration},
\textit{water solubility},
\textit{hydration free energy},
\textit{lipophilicity},
\textit{Tox21 toxicity outcomes},
\textit{clinical toxicity}, and
\textit{side-effect profile}.
The NoInst variant replaces \texttt{Encode this molecule for \{task\_name\}.}
with the generic instruction \texttt{Encode this molecule.} while keeping the
SMILES line and image token unchanged.

\paragraph{Structured target schema.} MolCAR-Structured target cards are
generated deterministically from ground-truth task labels and molecular
descriptors using the four-line schema:

\begin{AIbox}{MolCAR-Structured Target Template}
\small
Report family: \textit{\{family\_display\_name\}}\\
Task: \textit{\{task\_name\}}\\
Observed result: \textit{\{label\_sentence\}}\\
Molecular profile: \textit{\{descriptor\_sentence\}}
\end{AIbox}

No language model is involved at this stage: \texttt{label\_sentence} is
templated from the dataset's task label (e.g., a binary positive/negative
phrasing for classification, or a numeric value with units for regression),
and \texttt{descriptor\_sentence} is a fixed-format summary of MW, cLogP, HBD,
HBA, and formal charge computed from the molecule.

\paragraph{Natural target generation prompt.} MolCAR-Natural targets are
produced by prompting Intern-S1 (235B) once per (molecule, task) pair, then
re-anchoring the generated text with the trusted \texttt{Observed result} and
\texttt{Molecular profile} lines from the structured card. The user prompt
sent to the model is:

\begin{AIbox}{MolCAR-Natural Generation Prompt}
\small
\textbf{System.} Return only the final answer text. Do not output
chain-of-thought, analysis traces, or \texttt{<think>} tags.

\medskip
\textbf{User.} Write a 2--3 sentence scientist-facing retrieval note for this
molecule--task pair.\\
Requirements:
\begin{enumerate}[leftmargin=1.3em, itemsep=0em, topsep=0.2em]
\item Write naturally, like a scientific database annotation or research note.
\item Preserve task/context meaning; do not invent assays, outcomes, or claims.
\item You may include light mechanistic interpretation, but only as final
conclusions.
\item If you mention properties, do not enumerate full descriptor/value lists.
At most 1--2 key quantitative cues are acceptable when truly necessary.
\item Describe structural characteristics only when inferable from provided
context.
\item Do NOT output SMILES strings.
\item Do NOT output lines starting with ``Observed result:'' or
``Molecular profile:''.
\item Do NOT output reasoning traces, scratchpad text, or \texttt{<think>} blocks.
\item Return only the final note text.
\end{enumerate}

Dataset: \textit{\{source\_dataset\}}\\
Task context: \textit{\{context\_family\}} -- \textit{\{task\_name\}}\\
Key observation: \textit{\{observed\_summary\}}\\
Query context (SMILES hidden): \textit{\{query\_with\_smiles\_masked\}}

\medskip
Reference record (factual anchors):\\
\textit{\{structured\_target\_card\}}
\end{AIbox}

To prevent surface-form leakage, the SMILES inside the query is replaced with
\texttt{[hidden]} before being passed to the model.

\paragraph{Sampling parameters.} Generations use temperature 0.5, top-$p$ 1.0,
top-$k$ 50, and a 192-token cap, with thinking-mode disabled. The same
sampling configuration is used for the 4{,}092 evaluation pairs and the
10{,}844 MolCAR-Train pairs, so the structured and natural variants differ
only in target wording, not in molecule, task, or ground-truth outcome.

\paragraph{Anchor re-attachment.} After generation, the trusted
\texttt{Observed result} and \texttt{Molecular profile} lines from the
structured card are appended to the model output. This guarantees that ground-truth task
outcomes and physicochemical descriptors are preserved verbatim, while the
free-form portion of the document carries the natural-language framing.

\section{Implementation Details}
\label{app:implementation}

\subsection{Representation Extraction}

We follow the approach of VLM2Vec~\citep{jiang2024vlm2vec} and Qwen3-VL-Embedding~\cite{li2026qwen3} and extract the hidden state at the end-of-sequence (EOS) token position as the fixed-length embedding. The EOS position is appended to the end of the input sequence. We refer to this as EOS pooling. No additional pooling (mean, CLS) is applied.

\subsection{Molecule--Text Alignment Pretraining}

All alignment pretraining runs use bidirectional InfoNCE contrastive loss with a fixed temperature $\tau = 0.07$. The molecular query side
and the text target side are encoded by the same MLLM backbone through separate
forward passes. Representations are $\ell_2$-normalized before computing cosine
similarity. All experiments were conducted on 8 NVIDIA A100 GPUs.

All three alignment-pretraining corpora (MolTextNet, KnowMol-100k, ChEBI-20-MM) are split with a single shared script using a 98/1/1 random partition (train/valid/test, seed 42) at the molecule--text-pair level. Following the MolTextNet experimental protocol, we use the randomly sampled MolTextNet-300K subset rather than the full MolTextNet release~\cite{zhu2025moltextnet}. The Mixed condition then merges the three corpora and applies a global canonical-SMILES deduplication across all splits, so cross-corpus duplicates are removed rather than counted twice. Concrete corpus sizes after these steps are reported in Table~\ref{tab:alignment-corpora}. The motivation for the Mixed corpus is coverage rather than scale alone: it exposes the model to complementary forms of molecular language, including structural descriptions, functional group annotations, names, definitions, and physicochemical cues. The cross-modal retrieval probe in Section~\ref{sec:retrieval} of the main paper is evaluated on the standalone MolTextNet test split (2{,}970 pairs), not the Mixed test split, so the random R@1 baseline is $1/2970 \approx 0.034\%$.
\begin{table}[h]
\centering
\caption{Alignment-pretraining corpus sizes (molecule--text pairs). Pre-dedupe rows are each corpus's own 98/1/1 split. Mixed reflects the post-dedupe global merge across the three sources; merging drops 5{,}041 duplicate pairs at the canonical-SMILES level.}
\label{tab:alignment-corpora}
\begin{tabular}{lrrr}
\toprule
\textbf{Corpus} & \textbf{Train} & \textbf{Valid} & \textbf{Test} \\
\midrule
MolTextNet    & 290{,}962 & 2{,}969 & 2{,}970 \\
KnowMol-100k   &  94{,}968 &     969 &     970 \\
ChEBI-20-MM   &  30{,}630 &     312 &     314 \\
\midrule
Mixed  & 411{,}709 & 4{,}149 & 4{,}165 \\
\bottomrule
\end{tabular}
\end{table}
\paragraph{Optimization and adapter configuration.}
Across all three MLLM backbones, alignment pretraining shares the same training recipe: AdamW with weight decay $0.02$, bfloat16 mixed precision, LoRA on both language and vision components, learning rate $1\times 10^{-4}$ for the LoRA parameters, query/target max length $384/512$ tokens, image size $448$ px, and seed $42$. LoRA rank is backbone-specific: Qwen3.5-0.8B uses \(r=4\), whereas Intern-S1-mini and Qwen3-VL-8B use \(r=8\); in all cases \(\alpha=2r\). Training runs for up to $10$ epochs with early-stopping patience $3$ on validation loss.

\subsection{Property Prediction Adaptation}
\label{app:prop-adaptation}

For molecular property prediction (Section~\ref{sec:prediction}), each backbone is adapted with a lightweight task head on top of LoRA adapters using the same target groups and per-backbone rank/alpha as in alignment pretraining; for the MolTextNet and Mixed conditions, LoRA weights are inherited from the alignment-pretrained checkpoint, while Direct uses a freshly initialized LoRA. Constant across all runs: AdamW with weight decay $0.02$, bfloat16 mixed precision, max sequence length $512$ tokens, image size $448$ px, and up to $100$ epochs with early stopping on validation RMSE for regression and validation ROC-AUC for classification. The task head is a two-layer MLP with hidden size $512$; regression targets are standardized while classification targets are not. LoRA learning rate $\in\{5\times 10^{-5}, 1\times 10^{-4}\}$, head learning rate $\in\{1.5\times 10^{-4}, 3\times 10^{-4}, 5\times 10^{-4}, 1\times 10^{-3}\}$, and early-stopping patience $\in\{5, 10, 15\}$. Final results are averaged over $3$ random seeds $\{42, 43, 44\}$.

\subsection{Continued Alignment} 

Continued alignment uses the same contrastive objective as alignment pretraining and is initialized from an alignment-pretrained embedding model. For each MolCAR variant we run a parallel continued-alignment pass per backbone: the Structured continued-alignment checkpoint is trained on MolCAR-Train (Structured) and supplies the +Continued Alignment column of Table~\ref{tab:molcar_structured}, while the Natural continued-alignment checkpoint is trained on MolCAR-Train (Natural) and supplies the +Continued Alignment column of Table~\ref{tab:molcar_natural}; in each case the train and eval target distributions are matched in style. Both passes share the same recipe and differ only in the training corpus: a smaller LoRA learning rate of $5\times 10^{-5}$ to preserve the general molecular semantics acquired during alignment pretraining, and a fixed budget of $1000$ optimizer steps.

\section{Additional Results}
\subsection{MolCAR-Natural Results}\label{app:molcar_result}
MolCAR-Natural replaces structured target cards with scientist-facing notes synthesized by Intern-S1 (235B parameters), a large scientific-domain MLLM distinct from the MolEmb backbones evaluated in this paper. Each note is anchored to the same ground-truth outcome and molecular profile as the corresponding structured card, so the molecule, task, and observed result are held fixed while only the target text style varies. Table~\ref{tab:molcar_natural} reports the same Inst/NoInst/$\Delta$ evaluation protocol as Table~\ref{tab:molcar_structured}; the candidate pool contains the same 4{,}092 documents over 1{,}653 held-out molecules.

\begin{table}[h]
\centering
\caption{MolCAR-Natural results. We report Inst / NoInst / $\Delta$ for Base and +Continued Alignment. $\Delta$ is computed from unrounded scores before display; small discrepancies are due to rounding.}
\label{tab:molcar_natural}
\setlength{\tabcolsep}{4pt}
\begin{tabular}{ll|ccc|ccc}
\toprule
\textbf{Model} & \textbf{Metric} & \multicolumn{3}{c|}{\textbf{Base}} & \multicolumn{3}{c}{\textbf{+ Continued Alignment}} \\
\cmidrule(lr){3-5} \cmidrule(lr){6-8}
 & & \textbf{Inst (\%)} & \textbf{NoInst (\%)} & \textbf{$\Delta$ (pp)} & \textbf{Inst (\%)} & \textbf{NoInst (\%)} & \textbf{$\Delta$ (pp)} \\
\midrule
\multirow{5}{*}{Intern-S1-mini} & Mol.\ R@1 & 14.4 & 14.4 & \gdn{-0.1} & 22.9 & 18.1 & \gup{+4.8} \\
 & Doc.\ R@1 & 6.2 & 6.0 & \gup{+0.2} & 22.8 & 7.5 & \gup{+15.3} \\
 & Doc.\ MRR & 14.5 & 13.9 & \gup{+0.6} & 38.9 & 17.1 & \gup{+21.8} \\
 & Doc.\ R@5 & 21.4 & 20.1 & \gup{+1.3} & 57.6 & 25.0 & \gup{+32.6} \\
 & Context R@1 & 44.3 & 40.4 & \gup{+3.9} & \textbf{99.2} & 40.4 & \gup{\textbf{+58.8}} \\
\midrule
\multirow{5}{*}{Qwen3.5-0.8B} & Mol.\ R@1 & 9.7 & 9.4 & \gup{+0.3} & 21.8 & 16.9 & \gup{+4.9} \\
 & Doc.\ R@1 & 3.9 & 3.7 & \gup{+0.2} & 17.1 & 6.4 & \gup{+10.7} \\
 & Doc.\ MRR & 9.7 & 9.6 & \gup{+0.1} & 29.6 & 13.6 & \gup{+16.0} \\
 & Doc.\ R@5 & 14.0 & 14.1 & \gdn{-0.1} & 42.5 & 19.1 & \gup{+23.4} \\
 & Context R@1 & 40.6 & 40.4 & \gup{+0.2} & \textbf{72.0} & 40.4 & \gup{\textbf{+31.6}} \\
\bottomrule
\end{tabular}
\end{table}

\subsection{MolTextNet Retrieval Preservation after MolCAR Continued Alignment}
\label{app:moltextnet-preservation}

\begin{figure}[h]
\centering
\includegraphics[width=0.4\linewidth]{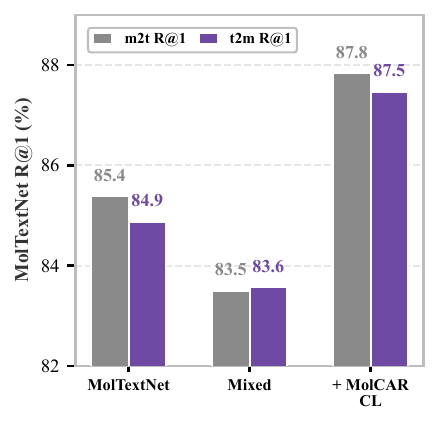}
\includegraphics[width=0.4\linewidth]{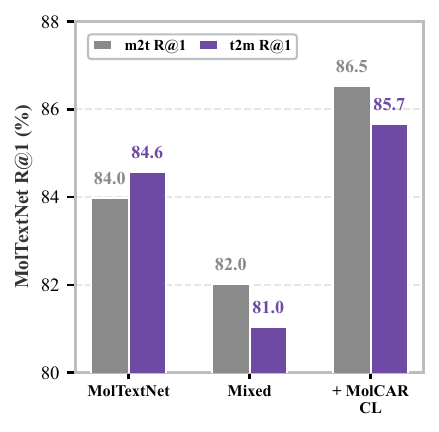}
\caption{MolTextNet retrieval remains strong after MolCAR continued alignment, for both Intern-S1-mini (left) and Qwen3.5-0.8B (right). Bars compare m2t/t2m R@1 on the held-out MolTextNet test split for the alignment-pretrained (Mixed) checkpoint and the same checkpoint after MolCAR continued alignment; the MolTextNet-only checkpoint is shown as a reference point. Here, CL denotes continued alignment.}
\label{fig:moltextnet-preservation}
\end{figure}

Figure~\ref{fig:moltextnet-preservation} compares MolTextNet m2t/t2m R@1 before and after MolCAR continued alignment for both backbones, with the MolTextNet-only checkpoint included as a reference point. Starting from the Mixed-aligned checkpoint, continued alignment on MolCAR-Train improves rather than degrades MolTextNet retrieval, indicating that the context-aware embedding capability discussed in Section~\ref{sec:context} is added on top of, rather than at the expense of, the general molecule--text alignment acquired during alignment pretraining.

\subsection{Controlled Instruction Probe t-SNE}
\label{app:tsne-prompt-probe}

To complement the retrieval-table evidence in Section~\ref{sec:context}, we visualize how task instructions reshape the embedding space under a controlled instruction-probe. Starting from a fixed molecule set, each unique molecule is expanded into all eight MolCAR task instructions, so changes in the resulting geometry reflect instruction conditioning rather than differences in sample composition. We then embed all (molecule, instruction) pairs through the embedding model and project them with t-SNE for visualization.

\begin{figure}[h]
\centering
\includegraphics[width=\textwidth]{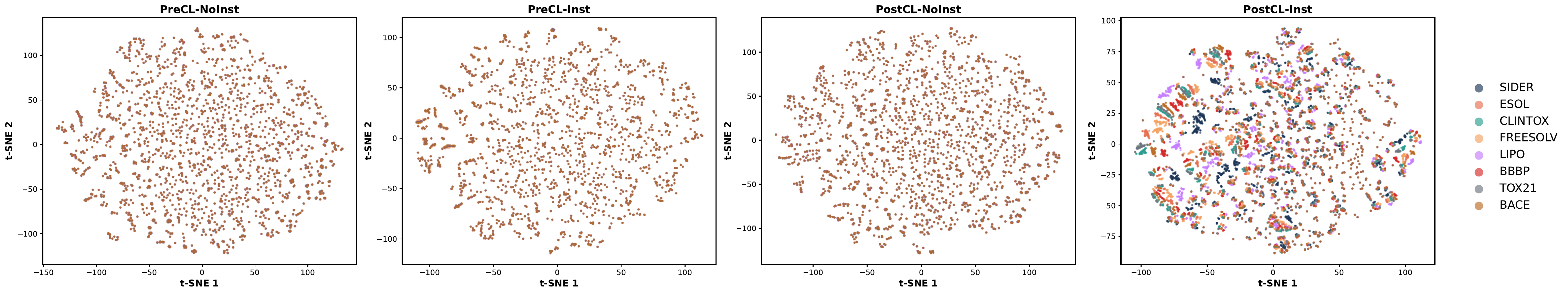}
\caption{Controlled MolCAR instruction-probe t-SNE on a fixed molecule set for Intern-S1-mini. Each unique molecule is expanded into all eight MolCAR task instructions, so changes in geometry reflect instruction conditioning rather than sample-composition drift. The first two panels show the model before continued alignment, and the last two panels show the model after continued alignment; each setting is evaluated with generic queries (NoInst) and task-instructed queries (Inst). Clear task-wise separation appears only after continued alignment and only when task instructions are present. PreCL and PostCL denote before and after continued alignment, respectively.}
\label{fig:molcar_prompt_probe}
\end{figure}

Figure~\ref{fig:molcar_prompt_probe} shows the result for Intern-S1-mini in four panels: before continued alignment with generic queries (NoInst) and with task-instructed queries (Inst), and after continued alignment under the same two query modes. Before continued alignment, the NoInst and Inst panels are visually similar, consistent with the close Inst/NoInst Context~R@1 scores reported in Table~\ref{tab:molcar_structured}. After continued alignment, the NoInst panel remains comparatively overlapped, while the Inst panel shows clear task-wise separation. The geometric view, therefore, tracks the quantitative result: continued alignment changes how the embedding model uses task instructions, not just how strongly it retrieves any single description.

\subsection{Multi-view Input Ablation}
\label{app:modality-ablation}

\begin{figure}
    \centering
    \includegraphics[width=1\linewidth]{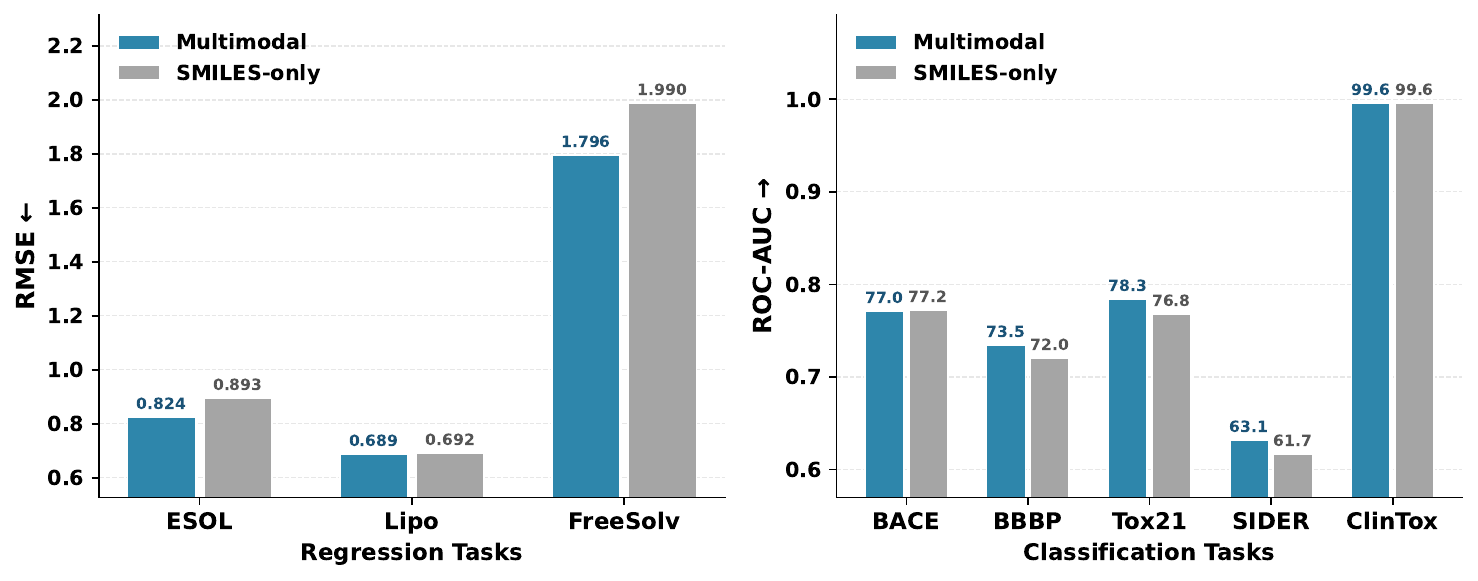}
    \caption{Multimodal profile versus SMILES-only input for Intern-S1-mini under direct adaptation. Using both the 2D depiction and canonical SMILES yields modest gains on most regression and classification tasks, supporting the value of the multi-view input interface.}
    \label{fig:modality}
\end{figure}

Figure~\ref{fig:modality} shows an ablation comparing the full multimodal profile (2D depiction + SMILES) against a SMILES-only baseline for Intern-S1-mini under direct adaptation. The 2D depiction contributes consistent though modest gains across most regression and classification tasks, supporting the multi-view input design in Section~\ref{sec:general-encoder}.

\clearpage
\section{Examples of Pretraining Corpora}
\label{app:alignment-corpus-examples}

  We show representative full-text supervision instances from the alignment pretraining corpora used in this work. Each example preserves the multimodal molecule-side query used during training, including an explicit image token, together with the full text-side target.

  \begin{AIbox}{MolTextNet Example}
  \small
  \textbf{Molecule query.}

  \texttt{<image>}\\
  Encode this molecule for reusable molecular understanding.\\
  Focus on its structure, substructures, functional groups, and physicochemical cues.\\
  SMILES: CCCCCCOc1cc(NC(=O)c2ccc(OC)cc2)ccc1N(C)S(C)(=O)=O

  \medskip
  \textbf{Text target.}

  presents a complex molecular architecture characterized by a molecular weight of 434.56 and a molecular formula of
  C$_{22}$H$_{30}$N$_2$O$_5$S. The compound exhibits a calculated logP value of 4.3, indicating moderate hydrophobicity, which
  may influence its membrane permeability and bioavailability. The presence of five hydrogen bond acceptors and one hydrogen
  bond donor suggests potential for significant molecular interactions with biological targets, while the polar surface area of
  84.94 supports solubility in aqueous environments. The compound does not violate Lipinski's rule of five, suggesting favorable
  drug-like properties. With 11 rotatable bonds, the compound's flexibility could enhance its ability to adopt various
  conformations, potentially optimizing binding interactions. The structural composition includes three aromatic rings and
  various functional groups such as an amide, ether, sulfonamide, and a secondary amine, which may contribute to the compound's
  overall bioactivity. However, the antiproliferative activity against human SKBR3 cells shows an IC50 value of 19980 nM,
  indicating a lack of significant activity. The synthetic accessibility scores, with an SCS score of 3.8192 and SAS score of
  2.0954, reflect a moderate level of synthetic complexity, suggesting that while the compound may be feasible to synthesize, it
  could present challenges due to its multiple functional groups and overall structural intricacies. The presence of a hexyloxy
  group may enhance lipophilicity, whereas the sulfonamide moiety could impart additional interactions with biological targets.
  \end{AIbox}

  \begin{AIbox}{KnowMol-100k Example}
  \small
  \textbf{Molecule query.}

  \texttt{<image>}\\
  Encode this molecule for reusable molecular understanding.\\
  Focus on its structure, substructures, functional groups, and physicochemical cues.\\
  SMILES: CCC(C)CC(C)C=C(C)C1OC(c2c(O)c(C3C(O)C(O)C(O)C3O)cn(C)c2=O)CC=C1C

  \medskip
  \textbf{Text target.}

  IUPAC name: 3-[6-[(E)-4,6-dimethyloct-2-en-2-yl]-5-methyl-3,6-dihydro-2H-pyran-2-yl]-4-hydroxy-1-methyl-5-(2,3,4,5-
  tetrahydroxycyclopentyl)pyridin-2-one. Functional groups: Alkyl, Alkenyl, Phenyl, Hydroxyl, Ether, Pyridyl. The molecule
  consists of several distinct substructures and functional groups. The main chain is an alkenyl chain with multiple methyl
  groups attached, forming a branched structure. This chain is connected to a dihydropyran ring, which is a six-membered ring
  containing one oxygen atom. Attached to the dihydropyran ring is a pyridine ring, which is a six-membered ring containing one
  nitrogen atom. The pyridine ring has a hydroxyl group and a methoxy group attached to it. Additionally, there is a
  tetrahydroxycyclopentyl group attached to the pyridine ring, which is a five-membered ring with four hydroxyl groups. The
  connections between these substructures are as follows: the alkenyl chain is connected to the dihydropyran ring, which is then
  connected to the pyridine ring. The tetrahydroxycyclopentyl group is attached to the pyridine ring. The molecule contains
  alkyl, alkenyl, hydroxyl, ether, and pyridyl functional groups. The molecule exhibits moderate polarity due to the presence of
  multiple hydroxyl groups and a pyridine ring, which introduce polar regions, although the overall structure is somewhat
  balanced by nonpolar alkyl chains. The presence of hydroxyl groups and the pyridine ring suggests that the molecule has both
  acidic and basic sites, with the hydroxyl groups contributing to acidity and the nitrogen in the pyridine ring contributing to
  basicity. The molecule is likely to be soluble in polar solvents like water due to the numerous hydroxyl groups, but the
  nonpolar alkyl chains may also allow some solubility in nonpolar solvents. Reactivity is influenced by the hydroxyl groups,
  which can participate in hydrogen bonding and nucleophilic reactions, and the double bond in the alkenyl chain, which can
  undergo addition reactions. The molecule has stereochemical complexity due to the presence of multiple chiral centers,
  particularly in the tetrahydroxycyclopentyl group, leading to potential enantiomers and diastereomers. Electrophilicity is
  enhanced by the electron-withdrawing effects of the carbonyl and hydroxyl groups, making certain sites more susceptible to
  nucleophilic attack.
  \end{AIbox}

  \begin{AIbox}{ChEBI-20-MM Example}
  \small
  \textbf{Molecule query.}

  \texttt{<image>}\\
  Encode this molecule for reusable molecular understanding.\\
  Focus on its structure, substructures, functional groups, and physicochemical cues.\\
  SMILES: COc1cc2c(=O)c(-c3ccc(O[C@@H]4O[C@H](C(=O)O)[C@@H](O)[C@H](O)[C@H]4O)cc3)\\coc2cc1O

  \medskip
  \textbf{Text target.}

  The molecule is a glycosyloxyisoflavone that is the glucuronide-conjugated form of the phytoestrogen glycitein. It is a 7-
  hydroxyisoflavone, a methoxyisoflavone, a glycosyloxyisoflavone, a beta-D-glucosiduronic acid and a monosaccharide derivative.
  It derives from a glycitein. IUPAC name: (2S,3S,4S,5R,6S)-3,4,5-trihydroxy-6-[4-(7-hydroxy-6-methoxy-4-oxochromen-3-
  yl)phenoxy]oxane-2-carboxylic acid. Physicochemical cues: XlogP 0.9 (moderate lipophilicity); polar surface area 172.0 A$^2$
  (high polar surface area).
  \end{AIbox}

\section{MolCAR Example}
\label{app:molcar-example}
This example makes the MolCAR construction concrete on a single held-out
molecule. The structured variant presents multiple candidate targets for the
same molecular profile under different task lenses, while the natural variant
replaces those same targets with scientist-facing notes anchored to the same
observed outcomes and descriptors. The instruction query and the generic
no-instruction query are both shown because MolCAR evaluates whether task
language, rather than molecule identity alone, routes the query to the correct
same-molecule target.
\begin{AIbox}{MolCAR-Structured Example}
  \small
  \textbf{Shared molecule.}

  \texttt{<image>}\\
  SMILES: BrC(Br)Br

  \medskip
  \textbf{Instruction query.}\\
  Encode this molecule for blood-brain barrier penetration.\\
  SMILES: BrC(Br)Br

  \medskip
  \textbf{No-instruction query.}\\
  Encode this molecule.\\
  SMILES: BrC(Br)Br

  \medskip
  \textbf{Candidate target cards.}

  \textbf{Doc. A}\\
  Report family: partition and barrier transport\\
  Task: blood-brain barrier penetration\\
  Observed result: This molecule crosses the blood-brain barrier.\\
  Molecular profile: MW 252.7; cLogP 2.45; HBD 0; HBA 0; formal charge 0

  \medskip
  \textbf{Doc. B}\\
  Report family: aqueous interaction\\
  Task: water solubility\\
  Observed result: Observed water solubility is -1.910 log mol/L.\\
  Molecular profile: MW 252.7; cLogP 2.45; HBD 0; HBA 0; formal charge 0

  \medskip
  \textbf{Doc. C}\\
  Report family: aqueous interaction\\
  Task: hydration free energy\\
  Observed result: Observed hydration free energy is -2.130 kcal/mol.\\
  Molecular profile: MW 252.7; cLogP 2.45; HBD 0; HBA 0; formal charge 0

  \medskip
  \textbf{Doc. D}\\
  Report family: safety liability\\
  Task: Tox21 toxicity outcomes\\
  Observed result: This molecule is positive in 0 of 12 observed Tox21 assay panel.\\
  Molecular profile: MW 252.7; cLogP 2.45; HBD 0; HBA 0; formal charge 0
  \end{AIbox}

\begin{AIbox}{MolCAR-Natural Example}
  \small
  \textbf{Shared molecule.}

  \texttt{<image>}\\
  SMILES: BrC(Br)Br

  \medskip
  \textbf{Instruction query.}\\
  Encode this molecule for blood-brain barrier penetration.\\
  SMILES: BrC(Br)Br

  \medskip
  \textbf{Candidate target notes.}

  \textbf{Doc. A}\\
  The molecule demonstrates effective blood-brain barrier penetration, suggesting favorable physicochemical properties for central nervous system targeting. With a molecular weight of 252.7 and moderate lipophilicity (cLogP 2.45), the compound lacks hydrogen bond donors or acceptors, potentially enhancing its passive diffusion across the BBB. These characteristics align with its observed ability to cross the blood-brain barrier.

  \medskip
  \textbf{Doc. B}\\
  The molecule exhibits moderate water solubility with an observed value of -1.910 log mol/L. Its physicochemical profile suggests a neutral, lipophilic character with a molecular weight of 252.7 and a calculated logP of 2.45, lacking hydrogen bond donors or acceptors. These properties likely contribute to its solubility behavior in aqueous environments.

  \medskip
  \textbf{Doc. C}\\
  The compound exhibits a measured hydration free energy of -2.130 kcal/mol, indicating moderate hydrophilic character despite its neutral charge and lack of hydrogen bond donors or acceptors. With a molecular weight of 252.7 and a cLogP of 2.45, the molecule demonstrates balanced lipophilicity and aqueous solubility properties. These features suggest limited polar interactions with water, consistent with its structural profile.

  \medskip
  \textbf{Doc. D}\\
  The molecule exhibits no positive responses across the 12 Tox21 assay panel, suggesting a low potential for toxicity in the tested endpoints. Its molecular profile, characterized by a moderate molecular weight (252.7) and balanced lipophilicity (cLogP 2.45), indicates favorable physicochemical properties that may contribute to its non-toxic profile. The absence of hydrogen bond donors and acceptors, along with a neutral formal charge, further supports its stability and reduced reactivity in biological systems.
  \end{AIbox}